%% file: main.tex
\documentclass[letterpaper, 10 pt, conference]{ieeeconf}  

\IEEEoverridecommandlockouts                              

\usepackage{amsmath,amsfonts,amssymb}
\usepackage{prettyref}
\usepackage{mathrsfs}
\usepackage{graphicx}
\usepackage{wrapfig}
\usepackage{subfig}
\let\labelindent\relax
\usepackage{enumitem}
\usepackage{MnSymbol}
\usepackage{multirow}
\usepackage{booktabs}
\usepackage{algorithm}
\usepackage[table]{xcolor}
\usepackage{cite}
\usepackage{caption}
\usepackage{ragged2e}
\usepackage{graphicx}

\title{\LARGE \bf
Physics-informed Neural Motion Planning on Constraint Manifolds
}

\author{Ruiqi Ni and Ahmed H. Qureshi
\thanks{The authors are with the Department of Computer Science, Purdue University, West Lafayette, IN 47907, USA \{ni117, ahqureshi\}@purdue.edu
}
}

\begin{document}

\maketitle
\thispagestyle{empty}
\pagestyle{empty}

\begin{abstract}

Constrained Motion Planning (CMP) aims to find a collision-free path between the given start and goal configurations on the kinematic constraint manifolds. These problems appear in various scenarios ranging from object manipulation to legged-robot locomotion. However, the zero-volume nature of manifolds makes the CMP problem challenging, and the state-of-the-art methods still take several seconds to find a path and require a computationally expansive path dataset for imitation learning. Recently, physics-informed motion planning methods have emerged that directly solve the Eikonal equation through neural networks for motion planning and do not require expert demonstrations for learning. Inspired by these approaches, we propose the first physics-informed CMP framework that solves the Eikonal equation on the constraint manifolds and trains neural function for CMP without expert data. Our results show that the proposed approach efficiently solves various CMP problems in both simulation and real-world, including object manipulation under orientation constraints and door opening with a high-dimensional 6-DOF robot manipulator. In these complex settings, our method exhibits high success rates and finds paths in sub-seconds, which is many times faster than the state-of-the-art CMP methods.
\end{abstract}

\input{01-introduction.tex}
\input{02-related.tex}

\input{03-problem.tex}
\input{04-method.tex}

\input{05-evaluation.tex}
\input{06-conclusion.tex}

\bibliography{reference}

\end{document}

%% file: 01-introduction.tex
\section{Introduction}
Constrained Motion Planning (CMP) is a challenging problem that aims to find a robot motion path between the given start and goal configurations such that the resulting path is collision-free and adheres to the given kinematic constraints. These kinematic constraints induce a thin manifold of zero volume inside robot configuration space, making finding a path solution challenging. Surprisingly, kinematic constraints appear in a variety of scenarios \cite{choset2005principles} ranging from object manipulation to solving robot locomotion tasks. Despite widespread applications, the existing solutions to solving CMP are limited by their computational inefficiency. 

The existing solutions to CMP broadly comprise optimization, sampling, and learning-based methods. The first set of methods defines constraints within a differentiable cost function and uses numerical optimization to find a path \cite{ratliff2009chomp,schulman2014motion,atkeson2015no,johnson2015team,dragan2011manipulation,bonalli2019trajectory}, making them susceptible to local minima. The second category is sampling-based motion planning (SMP) \cite{LaValle1998RapidlyExploringRT,latombe1998probabilistic, kuffner2000rrt,yao2005path,stilman2007task,berenson2011task,kim2016tangent,jaillet2017path,jaillet2013asymptotically,kingston2019exploring,bordalba2018randomized,englert2020sampling} which randomly sample the constraint manifolds and construct a graph for finding a path solution. Due to random sampling, these methods are computationally expensive and can take significant planning times to find a path. The modern techniques \cite{qureshi2020neural,qureshi2021constrained,lembono2021learning} leverage deep learning to approximate a sampler for generating samples on and nearby constraint manifolds for the given start and goal configurations. These samples are then given to SMP methods for planning. However, these methods require expert training data of motion paths on the manifolds for training. Such a dataset is usually gathered by running classical SMP methods, which leads to significant computational overload for training such learning-based approaches.

Recent advancements have led to Physics-Informed Neural Networks (PINN) that learn by directly solving Partial Differential Equations (PDE) representing a physical system \cite{raissi2019physics,smith2020eikonet}. The PINN has also been extended to solving robot motion planning problems under collision avoidance constraints \cite{ni2023ntfields,ni2023progressive}. These methods do not require training data comprising robot motion paths and, instead, learn to solve the Eikonal equation for path planning. Their results demonstrate that these methods outperform prior approaches in terms of computation times, provide high success rates, and scale to high-dimensional settings such as the 6 degree-of-freedom (DOF) robot arm. However, these PINN-based methods for motion planning have yet to be extended to solving CMP problems that induce kinematic constraints in addition to collision avoidance. 

Inspired by the abovementioned developments, this paper presents the first PINN-based method called constrained Neural Time Fields (C-NTFields) for solving CMP problems. Specifically, we extend the Eikonal equation formulation to incorporate kinematic constraints and demonstrate its application to training neural networks without expert training motion paths to solve CMP problems. We showcase our approach to tackling complex CMP tasks in simulations and the real world with a 6-DOF robot arm. These tasks involve challenges like handling objects with specific orientations and opening doors. Our results show that our method outperforms previous methods by a significant margin in terms of computational speed, path quality, and success rates. Furthermore, we also demonstrate that the data generation time for our method is a few minutes compared to the hours needed to gather path trajectories for traditional imitation learning-based neural planners.  

%% file: 02-related.tex
\section{Related Work}
In this section, we discuss the three major categories of existing CMP methods. The optimization-based methods \cite{ratliff2009chomp,schulman2014motion,atkeson2015no,johnson2015team,dragan2011manipulation} incorporate constraints into cost function and leverage numerical optimization for solving CMP tasks. However, these approaches weakly satisfy the kinematic constraints and are prone to local minima. A recent approach \cite{bonalli2019trajectory} solves trajectory optimization directly on the constraint manifold, but their application to high-dimensional CMP tasks has yet to be investigated. 

In contrast to optimization-based methods, the SMP methods \cite{LaValle1998RapidlyExploringRT,latombe1998probabilistic, kuffner2000rrt} have been widely investigated in a wide range of CMP problems. These methods rely on various sampling techniques to generate robot configuration samples on the constraint manifold and build a graph or tree for path planning. The existing sampling techniques within traditional SMPs rely on projection- or continuation-based operators \cite{berenson2009manipulation,jaillet2017path,kim2016tangent,kingston2019exploring}. The former projects the given sample to the manifold using iterative inverse kinematics-based numerical projection. The latter defines tangent spaces to approximate the manifold piecewise by forming an atlas. Hence, each manifold configuration can be linearly projected along the tangent space and mapped to the underlying manifold. These sampling approaches have been incorporated into various SMP methods, including PRMs and RRT-Connect. The modern SMP methods, such as RRT* and its variants, are usually not considered in CMP due to the high computational cost induced by their rewiring heuristic for optimal path planning. Even without optimal planning, the computational times with basic RRT-Connect and PRMs in solving CMP problems are still significantly high. 
 
Recently, a learning-based method called Constrained Motion Planning Networks X (CoMPNetX) \cite{qureshi2020neural,qureshi2021constrained} has been introduced for CMP. This method uses neural networks (NN) to generate robot configuration samples on and near manifolds for the underlying planner X (e.g., X= RRT-Connect). These NN are trained using imitation learning with the dataset of paths gathered from executing SMP planners in given CMP problems. Once trained, the NN generates samples in unseen but similar CMP problems as training data. Although this method exhibits fast planning speed at test time, the significant computational training load makes it less ideal as the total time to gather such training data surpasses the computational benefits at run time. Meanwhile, learning-based methods are also used for learning constraint manifolds \cite{sutanto2021learning} and for trajectory generation on constraint manifolds through reinforcement learning \cite{liu2022robot} and optimization method \cite{kicki2023fast}. However, these methods consider relatively simple scenarios with few obstacles, and their scalability to complex settings is yet to be explored. In this paper, we propose the PINN-based CMP method, which provides fast computational speed at test time in complex environments and does not require expensive training trajectories from classical planners for learning.

%% file: 03-problem.tex
\section{Background}
In this section, we present our problem definition along with notations used to describe the proposed approach. 
\subsection{Problem Definition}
Let the robot configuration space (C-space) and its surrounding environment be denoted as $\mathcal{Q} \in \mathbb{R}^m$ and $\mathcal{X}\in \mathbb{R}^d$, respectively. The $m$ and $d$ indicate the respective dimensions. The obstacles in the workspace are designated as $\mathcal{X}_{obs}$, leading to obstacle-free space defined as $\mathcal{X}_{free}= \mathcal{X}\setminus \mathcal{X}_{obs}$. This workspace obstacles map to robot C-space, yielding obstacle and obstacle-free space denoted as $\mathcal{Q}_{obs}$ and $\mathcal{Q}_{free}=\mathcal{Q}\setminus \mathcal{Q}_{obs}$, respectively. The objective of solving the robot motion planning problem is to find a path, $\sigma=\{q_0,\cdots,q_T\}$, in an obstacle-free configuration space that connects the given start, $q_0 \in \mathcal{Q}_{free}$, and goal, $q_T \in \mathcal{Q}_{free}$, i.e., $\sigma \subset \mathcal{Q}_{free}$. The CMP problem extends the standard motion planning problem by incorporating additional kinematic constraints. These constraints induce a thin manifold inside the robot C-space, which is denoted as $\mathcal{M} \subset \mathcal{Q}$. Like the C-space, the manifold also comprises the obstacle, $\mathcal{M}_{obs} \subset \mathcal{Q}_{obs}$, and obstacle-free space, $\mathcal{M}_{free} \subset \mathcal{Q}_{free}$. Finally, the objective of CMP is to find a robot motion path, $\sigma=\{q_0,\cdots,q_T\}$, between the given start, $q_0 \in \mathcal{M}_{free}$, and goal, $q_T \in \mathcal{M}_{free}$, such that $\sigma \subset \mathcal{M}_{free}$.

\subsection{Eikonal Equation Formulation}
{The Eikonal equation is a first-order non-linear PDE that finds the shortest arrival time $T(q_0,q_T)$ from $q_0$ to $q_T$ under the speed constraint $S(q_T)$ as follows:\vspace{-0.1in}}
\begin{equation}
\frac{1}{S(q_T)} = \|\nabla_{q_T} T(q_0,q_T)\|, 
\label{eikonal}
\end{equation}
where the $\nabla_{q_T} T(q_0,q_T)$ is the partial derivative of arrival time with respect to $q_T$. Recently, NTFields \cite{ni2023ntfields} extended the Eikonal formulation for path planning by formulating the arrival time in the following factorized form:
\begin{equation}
T(q_0,q_T)=\cfrac{\|q_0-q_T\|}{\tau(q_0,q_T)}   
\label{factorized}
\end{equation}
where the factorized time field is denoted by $\tau(q_0,q_T)$. 
{In the NTFields framework, given start and goal points as input, the neural network outputs the factorized time field $\tau$ between them,}
from which the speed is predicted using Equation \ref{eikonal}. The ground truth speed is calculated using the following predefined model: \vspace{-0.1in}
\begin{equation}
S^*(q)=\cfrac{s_{const}}{d_{max}}\times\mathrm{clip}(\boldsymbol{\mathrm{d}}_c(\boldsymbol{\mathrm{p}}(q),\mathcal{X}_{obs}), d_{min}, d_{max})  
\label{speed}
\end{equation}
where $d_{min}$ and $d_{max}$ are the predefined distance thresholds for the function $\boldsymbol{\mathrm{d}}_c(\cdot,\cdot)$ which returns the minimal distance between robot surface points $\boldsymbol{\mathrm{p}}(q)$ at configuration $q$ and the environment obstacles $\mathcal{X}_{obs}$. The robot surface points are computed using forward kinematics, and the $s_{const}$ is a predefined speed constant. {To find the shortest arrival time, the robot moves in the high-speed free space region and avoids the low-speed obstacle region.} Finally, the NTFields neural framework, predicting factorized arrival time, is trained using the isotropic loss function between predicted $S$ and ground truth $S^*$. The NTFields framework and its recent variant solve the motion planning problem under collision avoidance constraints. Our proposed framework in this paper can be seen as an application of NTFields to solving CMP problems without expert training path datasets. 

%% file: 04-method.tex
\section{Proposed Method}
This section formally presents our PINN framework for solving the CMP problem. Recall that the NTFields required robot configuration samples, the Eikonal equation formulation, speed definition, and strategies to leverage the predicted time field to compute the gradient steps for path generation. We discuss these individual components and their adaptions to solve CMP problems as follows:
\subsection{Manifold Configuration Sampling}
We formulate the kinematic constraints using implicitly defined Task Space Regions (TSR). A TSR comprises two SE(3) transformation matrices, i.e., \textbf{T}$^0_w$ and \textbf{T}$^w_e$, and a bound matrix \textbf{B}$^w$. The \textbf{T}$^0_w$ transforms from global coordinate to TSR frame $w$, also known as the object frame, whereas the \textbf{T}$^w_e$ provides an end-effector offset in frame $w$. The constraint manifold follows the equality constraint {by an implicit function} $f(\textbf{T}^0_w)=0$. Finally, the bound matrix, \textbf{B}$^w$, is a $6 \times 2$ matrix in TSR's coordinate frame $w$, and it formalizes the translational and rotational ranges of a TSR. These ranges are specifically chosen for different constraints. In our setting, we implicitly define this range by the distance of \textbf{T}$^0_w$ from the constraint manifolds, i.e., $\|f(\textbf{T}^0_w)\|<\delta$, where $\delta$ is a positive threshold, {and $\|f(\textbf{T}^0_w)\|$ represents the distance to the manifold which will be used for the following sampling and speed definition procedures}. 

Furthermore, we directly sample on the TSR $f(\textbf{T}^0_w)=0$ and use inverse kinematics to determine robot configurations on the manifold. Let's name them manifold configurations. We also add random perturbations to manifold configurations to gather off-manifold robot configurations and use the rejection strategy to gather samples within the implicit range. Using the sampling procedure described above, we generate a dataset of randomly sampled start and goal robot configurations, including cases of both on-manifold and off-manifold samples.
\subsection{ Expert Speed Model for Constraint Manifold}
The expert speed model defines the desired speed value of a given robot configuration, $q$, based on kinematic and collision avoidance constraints. The objective is to assign a maximum speed value to configuration samples on the collision-free constraint manifold, $\mathcal{M}_{free}$, and a lower speed value to collision and off-manifold samples. Let a function $\boldsymbol{\mathrm{d}}(q,\mathcal{M},\mathcal{X}_{obs}, \epsilon)$ determine the distance of a given configuration, $q$, from the collision-free constraint manifold, and $\epsilon$ be a predefined safety margin around obstacles. We define this function as:
\begin{equation}
\boldsymbol{\mathrm{d}}(q,\mathcal{X}_{obs}, \epsilon)=\max(\boldsymbol{\mathrm{d}}_\mathcal{M}(q), \epsilon-\boldsymbol{\mathrm{d}}_c(\boldsymbol{\mathrm{SDF}}(q), \mathcal{X}_{obs})) 
\label{distance}
\end{equation}
In the above formulation, the distance $\boldsymbol{\mathrm{d}}_\mathcal{M}$ measures the distance of a given configuration to the constraint manifold. We compute this distance following the $f(\cdot)$. Moreover, the function $\boldsymbol{\mathrm{d}}_c$ determines the minimum distance of a given robot configuration from the obstacles. To achieve $\boldsymbol{\mathrm{d}}_c$, we define the Signed Distance Functions (SDF) of the robot. These functions provide a value of zero, positive, or negative depending on whether a given obstacle point is on the robot surface, outside the robot, or inside the robot, respectively. Furthermore, we compute the robot's SDF at configuration $q$ using the function $\boldsymbol{\mathrm{SDF}}$. Finally, for these SDFs, the function $\boldsymbol{\mathrm{d}}_c$ returns the minimum distance of the robot from the obstacles. The two distance functions, $(\boldsymbol{\mathrm{d}}_\mathcal{M},\boldsymbol{\mathrm{d}}_c)$, and safety margin, $\epsilon$ are combined using the max operator.
The safety margin allows slow-speed maneuvering around obstacles, which is usually preferred over sharp turns offered by traditional planners. Moreover, in Eq. \ref{distance}, if the distance of collision surpasses the margin and creates a negative term $\epsilon-\boldsymbol{\mathrm{d}}_c$, the max operator will return the distance to the manifold since it is more important as the configuration is already far from the obstacle. Furthermore, note that the above distance function differs from the distance function used by NTFields, which only uses the distance from the obstacles to define their speed model. Next, we define our speed model based on distance function $\boldsymbol{\mathrm{d}}$ as follows:\vspace{-0.05in}
\begin{equation}
 S^*(q)=\exp(-\frac{\boldsymbol{\mathrm{d}}(q,\mathcal{X}_{obs},\epsilon)^2}{\lambda\epsilon^2}),   
\vspace{-0.05in}\end{equation}
where $\lambda \in \mathbb{R}^+$ is a predefined scaling factor. This speed model uses the negative exponential, which smoothly decays as the distance of the robot configuration from the collision-free manifold increases. In contrast, NTFields employs the clip function in their speed model to bind the minimum and maximum speed based on the robot's distance from the obstacles. Such a function cannot define the constraint manifold's speed model as the objective is to have a higher speed on the collision-free manifold and lower everywhere else, including the non-manifold, obstacle-free space. 

\subsection{Eikonal Equation Formulation}
The Eikonal equation is ill-posed, i.e., the solution of Eq. \ref{eikonal} around low-speed regions is not unique. Since kinematic constraint manifolds are of infinitesimal volume, this ill-posed nature of the Eikonal equation significantly affects PINN's performance. Recently, the progressive NTFields \cite{ni2023progressive} approach introduced a viscosity term based on Laplacian into Eikonal formulation, leading to a semi-linear elliptic PDE with a unique solution. \vspace{-0.1in}\begin{equation}
\frac{1}{S(q_T)} = \|\nabla_{q_T} T(q_0,q_T)\| + \eta \Delta_{q_T} T(q_0,q_T),
\label{vis}
\end{equation}
where $\eta \in \mathbb{R}$ is a scaling coefficient. The chain rule expansion of the above equation becomes:\\
$S(q_T)=$
\begin{equation}
\frac{1}{\eta \Delta_{q_T} \tau(q_0,q_T)+\sqrt{
\begin{aligned}
 [\tau^2(q_0,q_T)   - 2\tau(q_0,q_T)  (q_T-q_0)
\cdot\\ \nabla_{q_T} \tau(q_0,q_T)+\|q_0-q_T\|^2 
\times\\\|\nabla_{q_0} \tau(q_0,q_T)\|^2]/\tau^4(q_0,q_T) 
\end{aligned}
}}   
\label{viscoloss}
\end{equation}
We use the above formulation in our setting to overcome the challenges induced by the thin constraint manifold. Furthermore, similar to \cite{ni2023progressive}, we use $\Delta_{q_T} \tau(q_0,q_T)$ instead of $\Delta_{q_T} T(q_0,q_T)$ for computational simplification. Finally, the above equation has a unique solution and aids in training our PINN to represent constraint manifolds successfully.
\subsection{Neural Network Architecture}
Our neural architecture is similar to progressive NTFields methods \cite{ni2023progressive}. In summary, our framework can be formalized as follows:\vspace{-0.1in}
\begin{equation}
\tau(q_0,q_T)=g(\gamma(\mathcal{F}(q_0,Z))\bigotimes \gamma(\mathcal{F}(q_T,Z)))    
\end{equation}
In the above formulation, $\mathcal{F}$ is the Fourier transform-based environment and C-space encoder. It takes as an input the start and goal configurations, $(q_0,q_T)$, and pre-defined environment latent code $Z$, and outputs the high-frequency Fourier features:\vspace{-0.05in}
\begin{equation}
\begin{aligned}
    &\mathcal{F}(q_0,Z)=[\cos(2\pi Z^T q_0),\sin(2\pi Z^T q_0)]\\ &\mathcal{F}(q_T,Z)=[\cos(2\pi Z^T q_T),\sin(2\pi Z^T q_T)]
    \end{aligned}
    \label{rff}
\vspace{-0.05in}
\end{equation}
The output features, $(\mathcal{F}(q_0,Z),\mathcal{F}(q_T,Z))$, are then further embedded by a ResNet-style encoder, $\gamma$, with skip connections \cite{he2016deep}. Next, a symmetry operator, $\bigotimes$, combines the features using the $\max$ and $\min$ operators. For instance, some arbitrary inputs, $a$ and $b$, are combined as $a\bigotimes b=[\max(a,b), \min (a,b)]$ and $[\cdot]$ is a concetenation operator. The advantages of using a symmetric operator are discussed in \cite{ni2023ntfields}, which are that it maintains the symmetric property of arrival time, i.e., the arrival time from start to goal and from goal to start must be the same. Finally, the arrival time neural network, $g$, takes the symmetrically combined features and outputs the $\tau(q_0,q_T)$. This module also leverages the ResNet-style neural network with skip connections. The skip connection aids in the smooth gradient flow as highlighted in earlier works \cite{ni2023ntfields}. Finally, using the auto-differentiation, we compute the gradient $\nabla_{q_T} \tau(q_0,q_T)$ and the Laplacian $\Delta_{q_T} \tau(q_0,q_T)$ to determine $S(q_0)$ and $S(q_T)$, as described in Eq. \ref{viscoloss}.

\subsection{Training Procedure}
Given the start and goal configuration samples dataset generated on the manifolds and nearby, we train our above-mentioned neural network framework in an end-to-end manner. The NN module takes as an input the environment embedding $(Z)$, the start and goal configurations $(q_0,q_T)$, and outputs the factorized time $\tau(q_0,q_T)$. This factorized time is then used to predict the speed using Equation X. In addition, we also compute the ground truth speed using Equation 7. Finally, the NN can be trained by minimizing the following isotropic loss between the predicted and ground speed at the given configurations: \vspace{-0.05in}
\begin{equation}
    \begin{aligned}
    &S^*_{\beta(e)}(q_0)/S(q_0)+S(q_0)/S^*_{\beta(e)}(q_0)+\\
    &S^*_{\beta(e)}(q_T)/S(q_T)+S(q_T)/S^*_{\beta(e)}(q_T)-4
    \end{aligned}
    \label{trainloss}
\vspace{-0.05in}\end{equation}
Furthermore, we use the progressive speed scheduling approach to train our networks and prevent them from converging to incorrect local minima. The scheduling approach gradually scales down the ground truth speed from higher to lower value over the training epoch, $e$, using the parameter $\beta(e)$, i.e., $S^*_{\beta(e)}(q)=(1-\beta(e))+ \beta(e) S^*(q)$. This approach has already been demonstrated to overcome the complex loss landscape of physics-based objective functions and leads to better convergence in low-speed environments such as those with thin manifolds. Additionally, we employ a random batch buffer strategy to train our PINN method. This contrasts NTFields and P-NTFields, which process the entire dataset for each training epoch, leading to prolonged training times. However, our findings suggest that selecting a random, smaller data batch for each training epoch is a more efficient and effective approach. 
\subsection{Planning Procedure}
Once our NN modules are trained, we use the planning pipeline similar to the NTFields method. We begin by computing the factorized arrival time using NN, $\tau(q_0,q_T)$, required to travel from the starting point $q_0$ to the destination point $q_T$. Next, $\tau$ factorizes Eq. \ref{factorized} and \ref{eikonal} to compute $T(q_0,q_T)$ and speed fields $S(q_0), S(q_T)$. Finally, the start and goal configurations are bidirectionally and iteratively updated toward each other until the terminal limit is reached, i.e., $\|q_0-q_T\|<r_g$ to find a path, i.e.,
\begin{equation}
    \begin{aligned}
    q_0 &\gets q_0-\alpha S^2(q_0)\nabla^\perp_{q_0} T(q_0,q_T) \\
    q_T &\gets q_T-\alpha S^2(q_T)\nabla^\perp_{q_T} T(q_0,q_T)
    \end{aligned}
    \label{plan}
\end{equation}
where parameter $\alpha \in \mathbb{R}$ is a predefined step size and $r_g \in \mathbb{R}$ is predefined the goal region. 
Besides, in contrast to NTFields, where the gradient is only tangential, in the case of CMP, the gradient has two components, tangential and normal, due to the curved nature of the manifolds. Therefore, we select the tangential component $\nabla^\perp$ to move along the manifold for path planning.

%% file: 05-evaluation.tex
\section{Evaluation}

\begin{figure}[t]
\centering
\centering
    \includegraphics[width=0.14\textwidth]{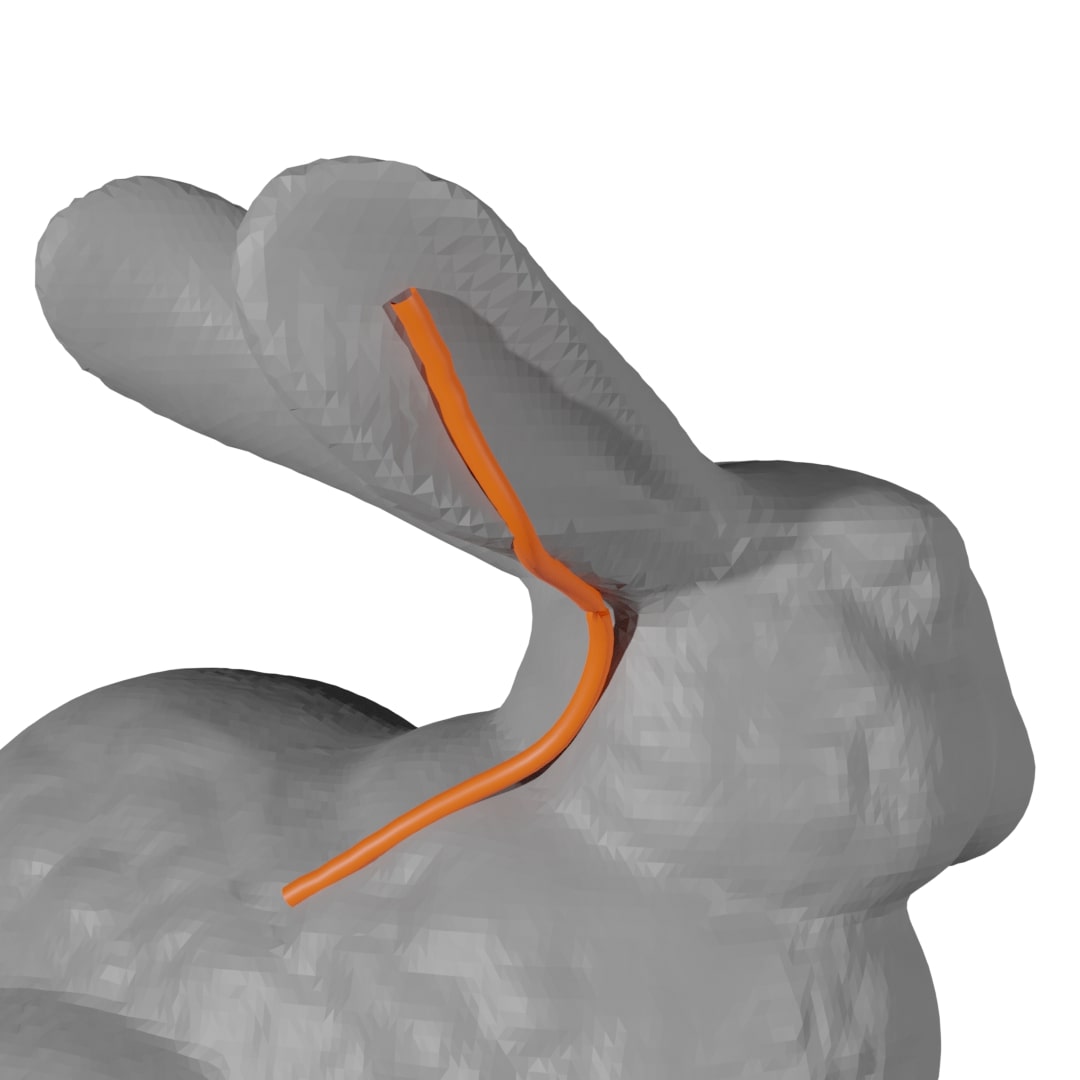}
    \includegraphics[width=0.14\textwidth]{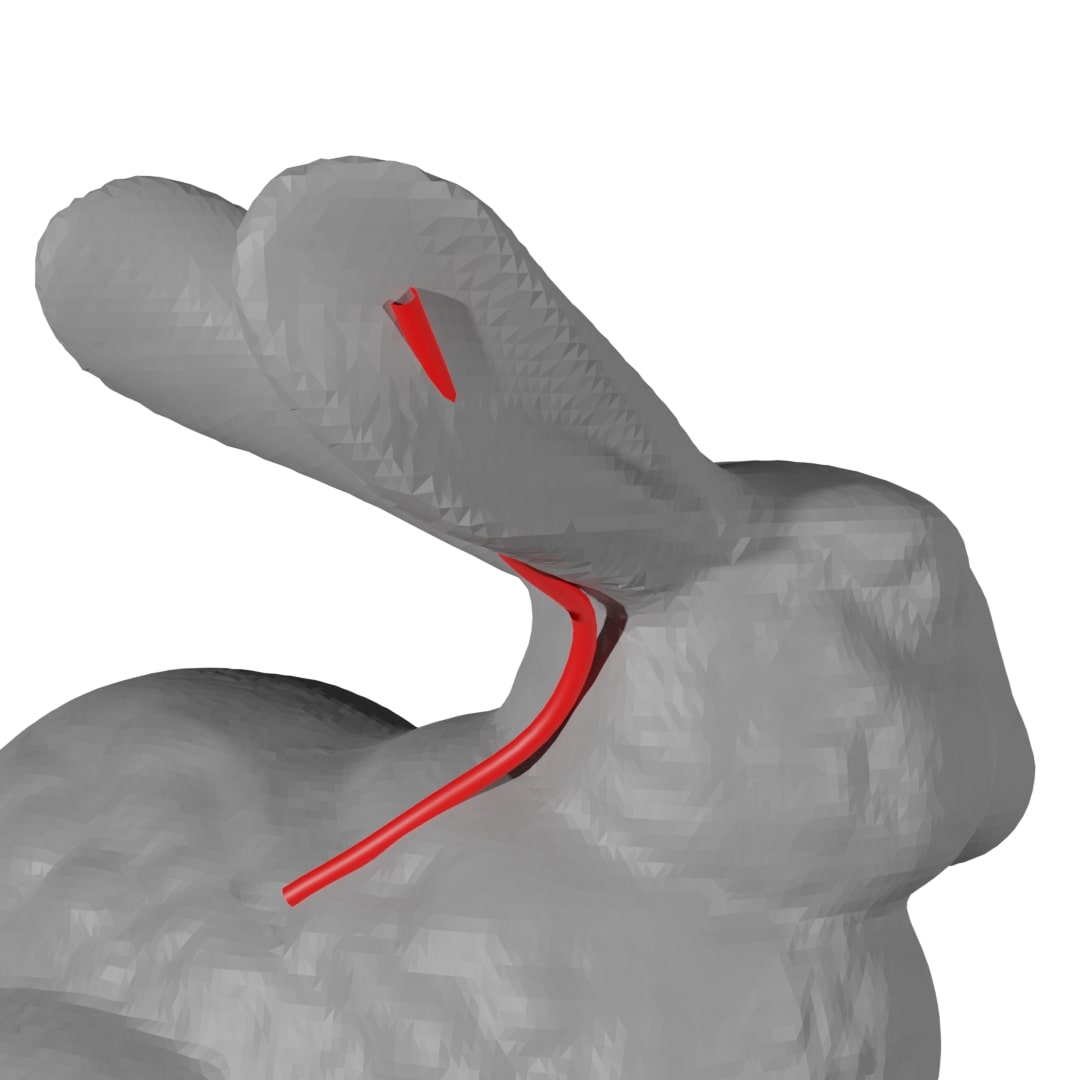}
    \includegraphics[width=0.14\textwidth]{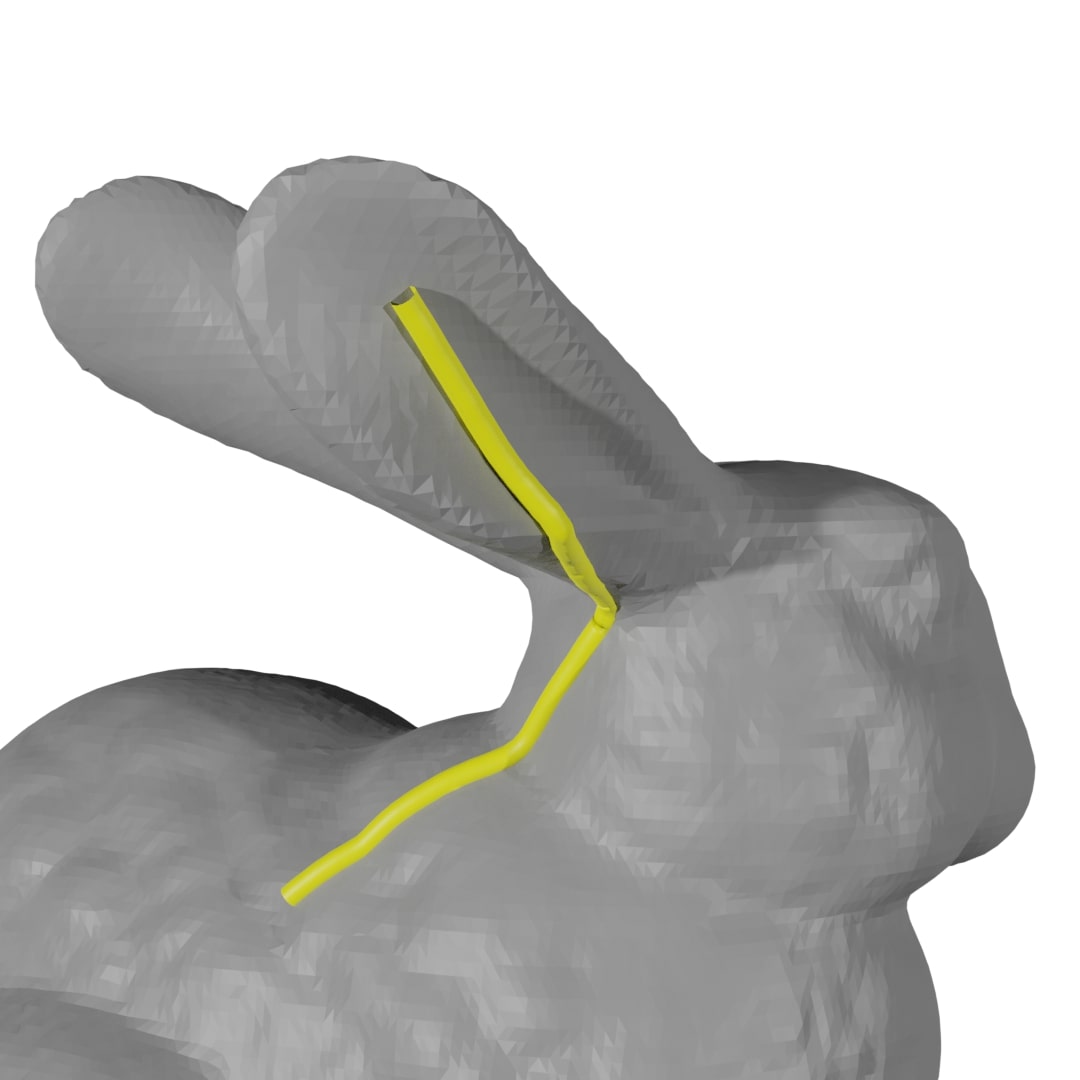}
\vspace{0.05in}
    \newline
\footnotesize
\setlength{\tabcolsep}{5pt}
\begin{tabular}{ccccc}
\hline
Bunny & time (sec)   &  length & {margin} & sr(\%)\\ 
\hline
Ours  & $ 0.06 \pm 0.05$ & $3.68 \pm 1.12$ & {$ 0.06 \pm 0.03 $} & 90 \\
        P-NTFields  & $ 0.05 \pm 0.05$ & $3.79 \pm 3.29$ & {$ 0.10 \pm 0.03$} & 79 \\
        HM  &  $ 0.05 \pm 0.00$ & $3.82 \pm 1.18 $ & {$0.00 \pm 0.00$} & 100 \\
\hline
\end{tabular}
\caption{\justifying From left to right images show the paths by our method (orange), P-NTFields (red), and HM (yellow). The statistical results are based on this environment's 100 different starts and goal pairs.}
\label{fig:abalation}
\vspace{-0.2in}
\end{figure}

\begin{figure}[t]
\centering
\includegraphics[width=0.18\textwidth,trim=0.0cm 2.0cm 0.0cm 0.4cm,clip]{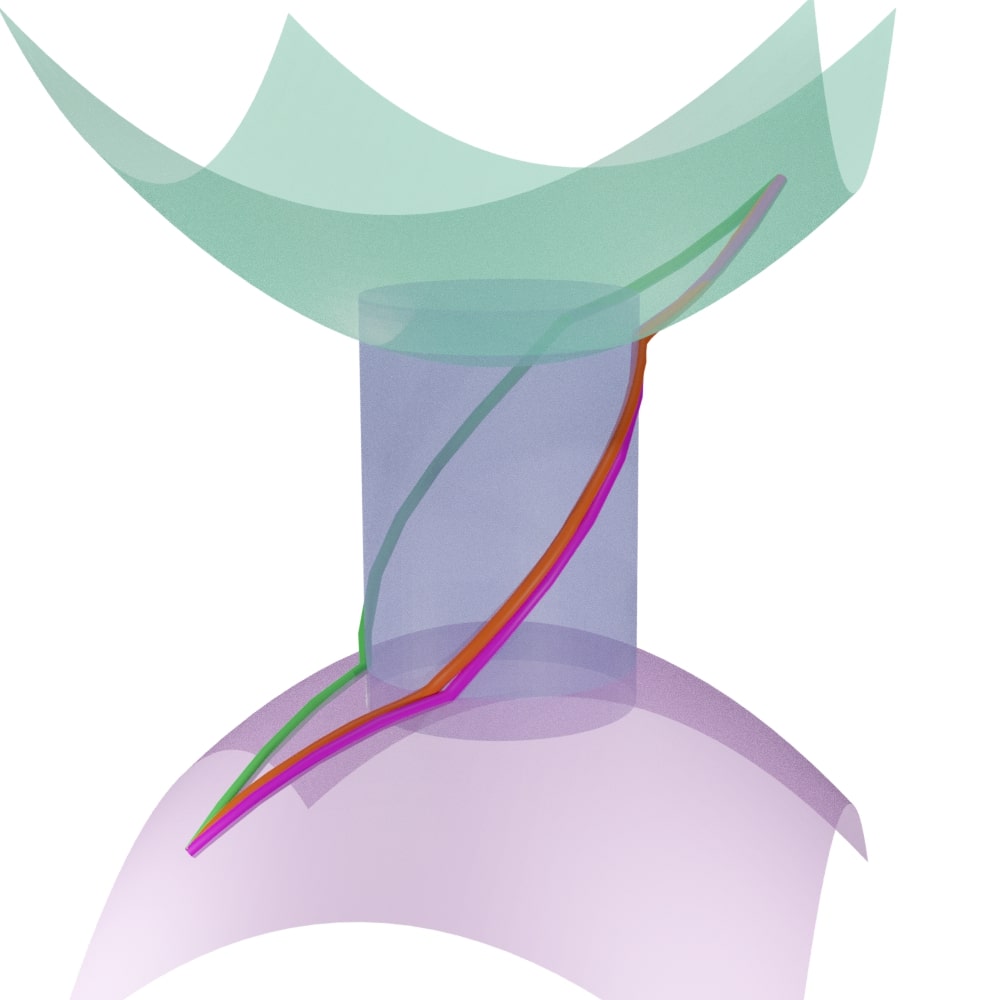}
\includegraphics[width=0.18\textwidth,trim=0.0cm 2.0cm 0.0cm 0.4cm,clip]{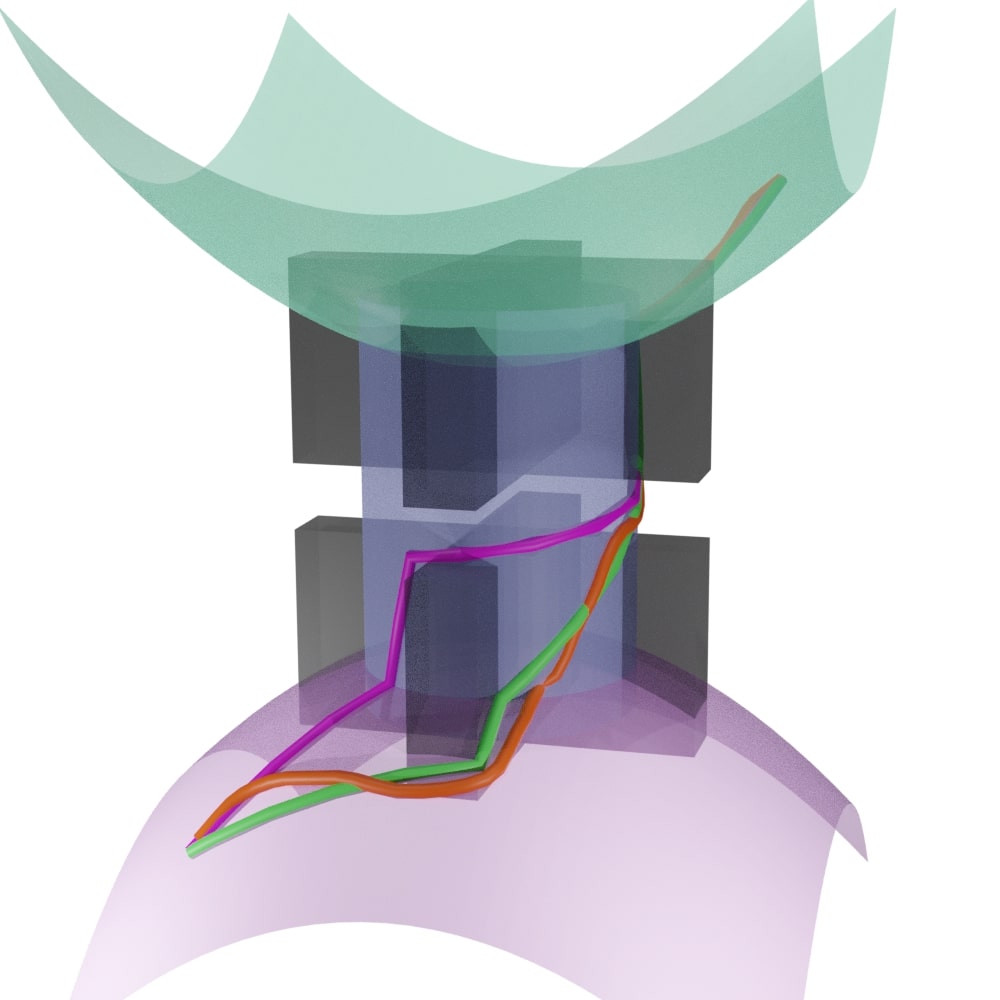}
\vspace{0.05in}
\footnotesize
\setlength{\tabcolsep}{5pt}
\begin{tabular}{ccccc}
\hline
 & time (sec)   &  length &  {margin} & sr(\%)\\ 
\hline
w/o obstacle\\
Ours  & $0.09\pm 0.00 $ & $14.32 \pm 0.00$ & $0.02 \pm 0.00$ & 100 \\

{ CBiRRT}   & {$1.59\pm 0.09$}  & {$14.70\pm 0.71$} & $0.00 \pm 0.00$ &100\\

{ CoMPNetX}   & {$0.38\pm 0.04$}  & {$14.79\pm 1.26$} & $0.00 \pm 0.00$ &100\\
\hline
w obstacle\\
Ours  & $0.12\pm 0.00 $ & $17.32\pm 0.00 $ & $0.11 \pm 0.00$ &100 \\

{ CBiRRT}   & {$1.10\pm 0.09$}  & {$16.66\pm 1.34$} & $0.00 \pm 0.00$ & 100\\

{ CoMPNetX}   & {$0.49\pm 0.12$}  & {$16.80\pm 1.81$} & $0.00 \pm 0.00$ &100\\
\hline
\end{tabular}
\caption{\justifying Without obstacles (left) and with obstacles (right) in 3D geometric constraint environments. The paths shown are from our method (orange), CBiRRT (pink), and CoMPNetX (green). The table shows the statistical results for different start and goal pairs in these settings.}
\label{fig:geometric}
\vspace{-0.25in}
\end{figure}

In this section, we assess the performance of our method through three sets of experiments. First, we employ ablation analysis to illustrate the efficacy of our novel speed model for representing constraint manifolds. Second, we conduct comparative experiments to evaluate our method against several state-of-the-art CMP baselines. Third, we analyze the data generation and training times of all learning-based methods. These experiments encompass the four problem settings: (1) A complex Bunny-shaped setting with 2D surface mesh in 3D space (Fig. \ref{fig:abalation}); (2) A geometrically constrained manifold in 3D space with and without obstacles (Fig. \ref{fig:geometric}); (3) A door-opening task with 6-DOF UR5e robot manipulator (Fig. \ref{fig:arm_real}); (4) An object manipulation task under orientation constraints with 6-DOF UR5e robot in intricate, narrow-passage cabinet environments (Fig. \ref{fig:arm_real}). For manipulator scenarios, we evaluate both the simulation and the real-world environments. Furthermore, we perform all experiments on a computing system with 3090 RTX GPU, Core i7 CPU, and 128GB RAM. Finally, the baseline methods and the evaluation metrics are summarized as follows: \\
\textit{Baselines:}
\begin{itemize} 
    \item \textbf{HM:} The heat method (HM), a diffusion-based method \cite{crane2013geodesics} that discretizes the given C-space manifold and solves the Eikonal equation for path planning. 
    \item \textbf{CBiRRT:} Two trees grow from the start and the goal towards each other with the projection-based method that adheres them to the constraint manifold \cite{berenson2009manipulation}.
    \item \textbf{CoMPNetX:} CoMPNetX provides informed samples for the underlying planner (e.g., RRT-Connect) to solve CMP. We chose Atlas as their constraint-adherence method due to its best performance in CoMPNetX experiments \cite{qureshi2021constrained}.
    \item \textbf{P-NTFields:} P-NTFields solve the Eikonal equation and do not require expert training data \cite{ni2023progressive}. Although P-NTFields do not consider manifold constraints, we still use its speed model for evaluation purposes. 
\end{itemize}
\noindent\textit{Evaluation Metrics:} 
\begin{itemize}
    \item \textbf{Time:} The time (in seconds) for a planner to find a valid path.
    \item \textbf{Length:} The path distance as the sum of Euclidean distance between its waypoints.
    \item \textbf{Margin:} The distance of waypoints to the constraint manifolds.
    \item \textbf{Success rate:} The percentage of valid paths found by a planner.
\end{itemize}

\subsection{Abalation Analysis}

This section analyzes our method's performance on a 2D surface mesh manifold (Bunny). We compare it with P-NTFields as ablation and HM as ground truth. Fig. \ref{fig:abalation} shows paths on the Bunny mesh of all methods, with the table providing their statistical comparison. From the paths in Fig. \ref{fig:abalation}, our method gets similar results to HM. However, P-NTFields penetrates the manifold. 

Furthermore, from the table in Fig. \ref{fig:abalation}, it can be seen that our method exhibits similar performance as ground truth method HM and has a higher success rate and lower margin to the manifold than P-NTField. This validates that our speed model design is suitable for constraint motion planning compared to the speed model definition in P-NTFields, which only applies to collision-avoidance constraints. Although HM works well for surface mesh examples, it requires discretization of the C-spaces and thus cannot generalize to higher dimensional robot settings. Therefore, we exclude HM and P-NTFields in the remainder experiment analysis.

\subsection{Comparison Analysis}

This section compares our method and other baselines on 2D geometric constraint manifold in 3D space and 6-DOF manipulator environments.

\begin{figure*}[ht!]
\centering
\includegraphics[width=0.95\textwidth]{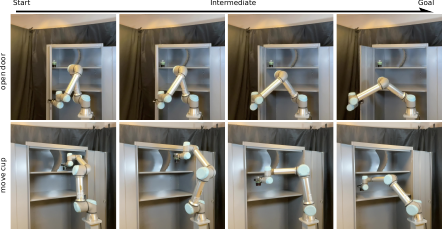}
    \caption{ \justifying Two different real-world manipulator cases: the first row shows the door opening task, whereas the second shows the manipulator moving an object from the cabinet's top shelf to the lower shelf by crossing two relatively thin obstacles.}
    \label{fig:arm_real}
\vspace{-0.1in}
\end{figure*}
\textbf{Geometric Constraints in 3D Space:}
Our geometric constraints setting includes three 2D manifolds in 3D space from \cite{englert2020sampling}. These manifolds are defined by parametric functions with paraboloid and cylinder shapes. We evaluate all methods on these manifolds with and without box obstacles. We use the same setting in \cite{englert2020sampling} and choose 10 random seeds for SMP methods, while our method is deterministic. Fig. \ref{fig:geometric} shows the paths of all presented methods with a table presenting the statistical comparison. All methods achieved 100\% success rate. However, it can be seen that the computational time of our method is about 3$\times$ and 10$\times$ faster than CoMPNetX and CBiRRT, respectively. Besides, our method does not require expensive data for learning, whereas CoMPNetX is trained using expert demonstration paths via imitation.

\textbf{Door Opening and Object Manipulation:}

These two tasks, defined by distinct manifolds, are solved through a 6-DOF UR5e Manipulator in both simulation and real-world settings. The door-opening task requires a robot to open the door from the current position to the open position. On the other hand, the object manipulation task imposes orientation constraints and requires the robot to maintain the object upright, i.e., without tilting, while moving it from a given start to the goal. We chose a challenging cabinet with narrow passages as our environment for these two tasks, which imposes significant motion planning challenges in terms of collision avoidance and manifold constraints. 

Table \ref{manipulator} compares our method, RRT-Connect, and CoMPNetX in the above-mentioned scenarios. For the object manipulation task, all methods have similar success rates and path lengths. However, our method is about 48$\times$ faster than CBiRRT and  15$\times$ faster than CoMPNetX in terms of computational times. For the door-opening task, since the constraint is relatively simple, all methods achieve similar results. Although our method exhibits a slight margin from the constraint manifold while others are strictly zero, this is because we do not use a hard constraint adherence approach on top of our framework. In contrast, other baselines, i.e., CoMPNext and CBiRRT, use an atlas and projection operator that generates samples on the manifold. In summary, it can be seen that our method outperforms both traditional and imitation learning-based methods on complex, high-dimensional scenarios by directly learning to solve the Eikonal equation without any expert path dataset.

Fig. \ref{fig:arm_real} shows our method's executions in real-world experiments. The environment was scanned via the RealSense sensor. The first row shows the snapshots of opening the door of the cabinet. This case took 0.06 seconds. The second row shows snapshots of moving a cup of cola across cabinet shelves: the manipulator moving from the cabinet's top shelf and crossing two relatively thin obstacles to another corner of the cabinet. This case took 0.15 seconds. We also provide a complete task execution video of opening a door and moving an object under orientation constraints between multiple starts and goals in our supplementary material.

\begin{table}[t]
\centering
\centering
\footnotesize
\setlength{\tabcolsep}{5pt}
\begin{tabular}{ccccc}
\hline
 & time (sec)   &  length &  {margin} & sr(\%)\\ 
\hline
move cup\\
Ours  & $0.14\pm 0.11 $ & $2.32 \pm 1.21$ & $0.04 \pm 0.03$ & 92 \\

{ CBiRRT}   & {$6.77\pm 6.40$}  & {$ 2.44 \pm 1.72 $} & $0.00 \pm 0.00$ &92\\

{ CoMPNetX}   & {$ 2.12 \pm 0.92$}  & {$2.43 \pm 1.56 $} & $0.00 \pm 0.00$ & 94\\
\hline

open door\\
Ours  & $0.05\pm 0.01 $ & $1.32 \pm 0.67$ & $0.05 \pm 0.05$ &100 \\

{ CBiRRT}   & {$0.06\pm 0.02$}  & {$1.30 \pm 0.62 $} & $0.00 \pm 0.00$ & 100\\

{ CoMPNetX}   & {$0.05\pm0.01 $}  & {$1.29 \pm 0.65 $} & $0.00 \pm 0.00$ & 100\\
\hline
\end{tabular}
\caption{\justifying The statistical results show 100 and 30 different starts and goal pairs for manipulating objects under orientation constraints and opening doors, respectively.}
\label{manipulator}
\vspace{-0.2in}
\end{table}

\subsection{Data Generation and Training Time Analysis}
Table \ref{datatime} shows the data generation and training times of our method, P-NTFields, and CoMPNetX. Our data generation time is significantly low, similar to P-NTFields, as we only need to compute robot samples and their distance to the manifold. In contrast, CoMPNetX requires expert trajectories from a classical planner and takes several hours in data generation for supervised learning. For the training time, P-NTFields take the longest time as it process the entire dataset for each training epoch. In comparison, our method training times are much lower than P-NTFields and somewhat similar to CoMPNetX due to our efficient mini-batches training during each epoch.

\begin{table}[t]
\centering
\centering
\footnotesize
\setlength{\tabcolsep}{5pt}
\begin{tabular}{cccc}
\hline
 Generation Time & Bunny   &  Geometric & Manipulator \\ 
\hline
Ours  & $3s $ & $3s $ & {$600s $}  \\
P-NTFields  & $3s $ & - & -  \\
CoMPNetX   & -  & $0.8h $ & $12h $  \\
\hline
\end{tabular}
\vspace{0.1in}

\centering
\footnotesize
\setlength{\tabcolsep}{5pt}
\begin{tabular}{cccc}
\hline
 Training Time & Bunny   &  Geometric & Manipulator \\ 
\hline
 Ours  & $2.5h$ & $2.5h$ & {$4.5h$}  \\
 P-NTFields  & $8h$ & - & -  \\
 CoMPNetX   & -  & $1h $ & $3h $ \\
\hline
\end{tabular}
\caption{Data generation and training times of our approach, P-NTFields, and CoMPNetX in different scenarios.}
\vspace{-0.2in}
\label{datatime}
\end{table}

%% file: 06-conclusion.tex
\section{Conclusions and Future Work} 
\label{sec:conclusion}
This paper presents the first physics-informed neural manipulation planning framework that finds paths on the kinematic constraint manifolds. Unlike the imitation learning-based method for CMP, which takes expert data, we show that our framework does not require expert demonstration path data and instead directly learns by solving the Eikonal equation. This leads to data generation times of a few seconds compared to hours for prior methods. Finally, our results also show that the proposed method is about 48$\times$ and 15$\times$ faster than classical and imitation-learning-based CMP methods in computation times in high-dimensional complex scenarios, including real-world object manipulation. In our future work, we aim to extend our approach to handle multimodal constraints, which often appear in legged robot locomotion tasks under contact dynamics. 

%% file: main.bbl
\begin{thebibliography}{10}
\providecommand{\url}[1]{#1}
\csname url@rmstyle\endcsname
\providecommand{\newblock}{\relax}
\providecommand{\bibinfo}[2]{#2}
\providecommand\BIBentrySTDinterwordspacing{\spaceskip=0pt\relax}
\providecommand\BIBentryALTinterwordstretchfactor{4}
\providecommand\BIBentryALTinterwordspacing{\spaceskip=\fontdimen2\font plus
\BIBentryALTinterwordstretchfactor\fontdimen3\font minus
  \fontdimen4\font\relax}
\providecommand\BIBforeignlanguage[2]{{%
\expandafter\ifx\csname l@#1\endcsname\relax
\typeout{** WARNING: IEEEtran.bst: No hyphenation pattern has been}%
\typeout{** loaded for the language `#1'. Using the pattern for}%
\typeout{** the default language instead.}%
\else
\language=\csname l@#1\endcsname
\fi
#2}}

\bibitem{atkeson2015no}
C.~G. Atkeson, B.~P.~W. Babu, N.~Banerjee, D.~Berenson, C.~P. Bove, X.~Cui,
  M.~DeDonato, R.~Du, S.~Feng, P.~Franklin, \emph{et~al.}, ``No falls, no
  resets: Reliable humanoid behavior in the darpa robotics challenge,'' in
  \emph{2015 IEEE-RAS 15th International Conference on Humanoid Robots
  (Humanoids)}.\hskip 1em plus 0.5em minus 0.4em\relax IEEE, 2015, pp.
  623--630.

\bibitem{berenson2011task}
D.~Berenson, S.~Srinivasa, and J.~Kuffner, ``Task space regions: A framework
  for pose-constrained manipulation planning,'' \emph{The International Journal
  of Robotics Research}, vol.~30, no.~12, pp. 1435--1460, 2011.

\bibitem{berenson2009manipulation}
D.~Berenson, S.~S. Srinivasa, D.~Ferguson, and J.~J. Kuffner, ``Manipulation
  planning on constraint manifolds,'' in \emph{2009 IEEE international
  conference on robotics and automation}.\hskip 1em plus 0.5em minus
  0.4em\relax IEEE, 2009, pp. 625--632.

\bibitem{bonalli2019trajectory}
R.~Bonalli, A.~Bylard, A.~Cauligi, T.~Lew, and M.~Pavone, ``Trajectory
  optimization on manifolds: A theoretically-guaranteed embedded sequential
  convex programming approach,'' \emph{arXiv preprint arXiv:1905.07654}, 2019.

\bibitem{bordalba2018randomized}
R.~Bordalba, L.~Ros, and J.~M. Porta, ``Randomized kinodynamic planning for
  constrained systems,'' in \emph{2018 IEEE international conference on
  robotics and automation (ICRA)}.\hskip 1em plus 0.5em minus 0.4em\relax IEEE,
  2018, pp. 7079--7086.

\bibitem{choset2005principles}
H.~Choset, K.~M. Lynch, S.~Hutchinson, G.~A. Kantor, and W.~Burgard,
  \emph{Principles of robot motion: theory, algorithms, and
  implementations}.\hskip 1em plus 0.5em minus 0.4em\relax MIT press, 2005.

\bibitem{crane2013geodesics}
K.~Crane, C.~Weischedel, and M.~Wardetzky, ``Geodesics in heat: A new approach
  to computing distance based on heat flow,'' \emph{ACM Transactions on
  Graphics (TOG)}, vol.~32, no.~5, pp. 1--11, 2013.

\bibitem{dragan2011manipulation}
A.~D. Dragan, N.~D. Ratliff, and S.~S. Srinivasa, ``Manipulation planning with
  goal sets using constrained trajectory optimization,'' in \emph{2011 IEEE
  International Conference on Robotics and Automation}.\hskip 1em plus 0.5em
  minus 0.4em\relax IEEE, 2011, pp. 4582--4588.

\bibitem{englert2020sampling}
P.~Englert, I.~M.~R. Fern{\'a}ndez, R.~K. Ramachandran, and G.~S. Sukhatme,
  ``Sampling-based motion planning on sequenced manifolds,'' \emph{arXiv
  preprint arXiv:2006.02027}, 2020.

\bibitem{he2016deep}
K.~He, X.~Zhang, S.~Ren, and J.~Sun, ``Deep residual learning for image
  recognition,'' in \emph{Proceedings of the IEEE conference on computer vision
  and pattern recognition}, 2016, pp. 770--778.

\bibitem{jaillet2017path}
L.~Jaillet and J.~M. Porta, ``Path planning with loop closure constraints using
  an atlas-based rrt,'' in \emph{Robotics Research: The 15th International
  Symposium ISRR}.\hskip 1em plus 0.5em minus 0.4em\relax Springer, 2017, pp.
  345--362.

\bibitem{jaillet2013asymptotically}
------, ``Asymptotically-optimal path planning on manifolds,'' in
  \emph{Robotics: Science and Systems}, vol.~8, 2013, p. 145.

\bibitem{johnson2015team}
M.~Johnson, B.~Shrewsbury, S.~Bertrand, T.~Wu, D.~Duran, M.~Floyd, P.~Abeles,
  D.~Stephen, N.~Mertins, A.~Lesman, \emph{et~al.}, ``Team ihmc's lessons
  learned from the darpa robotics challenge trials,'' \emph{Journal of Field
  Robotics}, vol.~32, no.~2, pp. 192--208, 2015.

\bibitem{kicki2023fast}
P.~Kicki, P.~Liu, D.~Tateo, H.~Bou-Ammar, K.~Walas, P.~Skrzypczy{\'n}ski, and
  J.~Peters, ``Fast kinodynamic planning on the constraint manifold with deep
  neural networks,'' \emph{arXiv preprint arXiv:2301.04330}, 2023.

\bibitem{kim2016tangent}
B.~Kim, T.~T. Um, C.~Suh, and F.~C. Park, ``Tangent bundle rrt: A randomized
  algorithm for constrained motion planning,'' \emph{Robotica}, vol.~34, no.~1,
  pp. 202--225, 2016.

\bibitem{kingston2019exploring}
Z.~Kingston, M.~Moll, and L.~E. Kavraki, ``Exploring implicit spaces for
  constrained sampling-based planning,'' \emph{The International Journal of
  Robotics Research}, vol.~38, no. 10-11, pp. 1151--1178, 2019.

\bibitem{kuffner2000rrt}
J.~J. Kuffner and S.~M. LaValle, ``Rrt-connect: An efficient approach to
  single-query path planning,'' in \emph{Proceedings 2000 ICRA. Millennium
  Conference. IEEE International Conference on Robotics and Automation.
  Symposia Proceedings (Cat. No. 00CH37065)}, vol.~2.\hskip 1em plus 0.5em
  minus 0.4em\relax IEEE, 2000, pp. 995--1001.

\bibitem{latombe1998probabilistic}
L.~E. K. J.-C. Latombe, ``Probabilistic roadmaps for robot path planning,''
  \emph{Pratical motion planning in robotics: current aproaches and future
  challenges}, pp. 33--53, 1998.

\bibitem{LaValle1998RapidlyExploringRT}
\BIBentryALTinterwordspacing
S.~M. LaValle, ``Rapidly-exploring random trees: A new tool for path
  planning,'' 1998. [Online]. Available:
  \url{https://api.semanticscholar.org/CorpusID:14744621}
\BIBentrySTDinterwordspacing

\bibitem{lembono2021learning}
T.~S. Lembono, E.~Pignat, J.~Jankowski, and S.~Calinon, ``Learning constrained
  distributions of robot configurations with generative adversarial network,''
  \emph{IEEE Robotics and Automation Letters}, vol.~6, no.~2, pp. 4233--4240,
  2021.

\bibitem{liu2022robot}
P.~Liu, D.~Tateo, H.~B. Ammar, and J.~Peters, ``Robot reinforcement learning on
  the constraint manifold,'' in \emph{Conference on Robot Learning}.\hskip 1em
  plus 0.5em minus 0.4em\relax PMLR, 2022, pp. 1357--1366.

\bibitem{ni2023ntfields}
R.~Ni and A.~H. Qureshi, ``{NTF}ields: Neural time fields for physics-informed
  robot motion planning,'' in \emph{The Eleventh International Conference on
  Learning Representations}, 2023.

\bibitem{ni2023progressive}
------, ``Progressive learning for physics-informed neural motion planning,''
  \emph{arXiv preprint arXiv:2306.00616}, 2023.

\bibitem{qureshi2020neural}
A.~H. Qureshi, J.~Dong, A.~Choe, and M.~C. Yip, ``Neural manipulation planning
  on constraint manifolds,'' \emph{IEEE Robotics and Automation Letters},
  vol.~5, no.~4, pp. 6089--6096, 2020.

\bibitem{qureshi2021constrained}
A.~H. Qureshi, J.~Dong, A.~Baig, and M.~C. Yip, ``Constrained motion planning
  networks x,'' \emph{IEEE Transactions on Robotics}, vol.~38, no.~2, pp.
  868--886, 2021.

\bibitem{raissi2019physics}
M.~Raissi, P.~Perdikaris, and G.~E. Karniadakis, ``Physics-informed neural
  networks: A deep learning framework for solving forward and inverse problems
  involving nonlinear partial differential equations,'' \emph{Journal of
  Computational physics}, vol. 378, pp. 686--707, 2019.

\bibitem{ratliff2009chomp}
N.~Ratliff, M.~Zucker, J.~A. Bagnell, and S.~Srinivasa, ``Chomp: Gradient
  optimization techniques for efficient motion planning,'' in \emph{2009 IEEE
  international conference on robotics and automation}.\hskip 1em plus 0.5em
  minus 0.4em\relax IEEE, 2009, pp. 489--494.

\bibitem{schulman2014motion}
J.~Schulman, Y.~Duan, J.~Ho, A.~Lee, I.~Awwal, H.~Bradlow, J.~Pan, S.~Patil,
  K.~Goldberg, and P.~Abbeel, ``Motion planning with sequential convex
  optimization and convex collision checking,'' \emph{The International Journal
  of Robotics Research}, vol.~33, no.~9, pp. 1251--1270, 2014.

\bibitem{smith2020eikonet}
J.~D. Smith, K.~Azizzadenesheli, and Z.~E. Ross, ``Eikonet: Solving the eikonal
  equation with deep neural networks,'' \emph{IEEE Transactions on Geoscience
  and Remote Sensing}, vol.~59, no.~12, pp. 10\,685--10\,696, 2020.

\bibitem{stilman2007task}
M.~Stilman, ``Task constrained motion planning in robot joint space,'' in
  \emph{2007 IEEE/RSJ International Conference on Intelligent Robots and
  Systems}.\hskip 1em plus 0.5em minus 0.4em\relax IEEE, 2007, pp. 3074--3081.

\bibitem{sutanto2021learning}
G.~Sutanto, I.~R. Fern{\'a}ndez, P.~Englert, R.~K. Ramachandran, and
  G.~Sukhatme, ``Learning equality constraints for motion planning on
  manifolds,'' in \emph{Conference on Robot Learning}.\hskip 1em plus 0.5em
  minus 0.4em\relax PMLR, 2021, pp. 2292--2305.

\bibitem{yao2005path}
Z.~Yao and K.~Gupta, ``Path planning with general end-effector constraints:
  Using task space to guide configuration space search,'' in \emph{2005
  IEEE/RSJ International Conference on Intelligent Robots and Systems}.\hskip
  1em plus 0.5em minus 0.4em\relax IEEE, 2005, pp. 1875--1880.

\end{thebibliography}
